\documentclass[lettersize,journal]{IEEEtran}
\usepackage{amsmath,amsfonts}
\usepackage{algorithmic}
\usepackage{algorithm}
\usepackage{array}
\usepackage[caption=false,font=normalsize,labelfont=sf,textfont=sf]{subfig}
\usepackage{textcomp}
\usepackage{stfloats}
\usepackage{url}
\usepackage{verbatim}
\usepackage{graphicx}
\usepackage{cite}
\usepackage{balance}
\immediate\write18{bibtex \jobname}
\begin{document}

\title{TauGenNet: Plasma-Driven Tau PET Image Synthesis via Text-Guided 3D Diffusion Models}

\author{Yuxin Gong, Se-in Jang, Wei Shao, Yi Su, Kuang Gong, for the Alzheimer’s Disease Neuroimaging Initiative (ADNI)
\thanks{This work was supported by NIH grants R01EB034692 and
		R01AG078250. (Corresponding author: Kuang Gong)}\thanks{Y. Gong and K. Gong are with the J. Crayton Pruitt Family Department of Biomedical Engineering, University of Florida, Gainesville, FL, 32611, USA. (e-mail: gongyuxin@ufl.edu,KGong@bme.ufl.edu). }
    \thanks{S. Jang is with the Department of Radiology \& Biomedical Imaging, Yale University (e-mail:sein.jang@yale.edu). }
    \thanks{W. Shao is with the Department of Medicine, University of Florida, Gainesville, FL, 32611, USA. (e-mail: weishao@ufl.edu). }
    \thanks{Y. Su is with Banner Alzheimer’s Institute, Phoenix, AZ, 85006, USA. (e-mail: Yi.Su@bannerhealth.com). }
}

% The paper headers
\markboth{Journal of \LaTeX\ Class Files,~Vol.~14, No.~8, August~2021}%
{Shell \MakeLowercase{\textit{et al.}}: A Sample Article Using IEEEtran.cls for IEEE Journals}

\maketitle

\begin{abstract}
Accurate quantification of tau pathology via tau positron emission tomography (PET) scan is crucial for diagnosing and monitoring Alzheimer’s disease (AD). However, the high cost and limited availability of tau PET restrict its widespread use. In contrast, structural magnetic resonance imaging (MRI) and plasma-based biomarkers provide non-invasive and widely available complementary information related to brain anatomy and disease progression. In this work, we propose a text-guided 3D diffusion model for 3D tau PET image synthesis, leveraging multimodal conditions from both structural MRI and plasma measurement. Specifically, the textual prompt is from the plasma p-tau217 measurement, which is a key indicator of AD progression, while MRI provides anatomical structure constraints. The proposed framework is trained and evaluated using clinical AV1451 tau PET data from the Alzheimer’s Disease Neuroimaging Initiative (ADNI) database. Experimental results demonstrate that our approach can generate realistic, clinically meaningful 3D tau PET across a range of disease stages. The proposed framework can help perform tau PET data augmentation under different settings, provide a non-invasive, cost-effective alternative for visualizing tau pathology, and support the simulation of disease progression under varying plasma biomarker levels and cognitive conditions.

\end{abstract}


\section{Introduction}
\IEEEPARstart{T} 
au PET imaging plays an important role in studying Alzheimer’s disease (AD) and related neurodegenerative disorders\cite{johnson2016tau}. By visualizing the spatial distribution of tau neurofibrillary tangles (NFT) {\it{in vivo}}, tau PET enables precise staging of disease progression and offers critical insights into the patterns of pathological spread. According to Braak staging, tau NFTs first emerge in the transentorhinal cortex, gradually spreading to the entorhinal cortex, hippocampus, and other cortical regions \cite{braak1991neuropathological}. Recent breakthroughs in radiolabeled tracers \cite{chien2013early,hostetler2016preclinical,kroth2019discovery} have positioned tau PET as the only modality capable of mapping tau spatial distribution {\it{in vivo}}. Studies have shown that tau PET aligns with Braak staging \cite{schwarz2016regional}, correlates strongly with cognitive measures \cite{pontecorvo2017relationships}, and can capture clinical and neuroanatomical variability of AD \cite{ossenkoppele2016tau}. These capabilities make tau PET essential for early AD diagnosis and progression tracking. 

% Tracers such as AV1451 have demonstrated high sensitivity and specificity in detecting tau accumulation in key brain regions affected by AD\cite{marquie2017f}. 
Although tau PET enables the {\it{in vivo}} visualization of tau pathology and is considered the gold standard for assessing AD progression, it is expensive and not easily accessible. Tau PET is currently limited to research studies and clinical trials, with only a few publicly available datasets \cite{mueller2005alzheimer,lamontagne2019oasis,dagley2017harvard}. In cross-sectional studies, missing tau PET data presents a substantial challenge, particularly in multi-cohort studies where the absence of such imaging hinders comparative analyses\cite{estarellas2024multimodal}. Additionally, machine learning models relying on tau PET require large, well-balanced datasets to ensure generalizability and robustness\cite{karlsson2025machine,yoon2019effect}. However, the limited availability of tau PET scans, coupled with inherent class imbalances in disease cohorts, compromises model performance. To address both missing data and small dataset limitations, synthetic tau PET generation based on multimodal measures is needed. Furthermore, by varying input conditions, the generated tau PET images can help explore how tau pathology relates to demographic factors, clinical measures, neuroimaging findings, and plasma biomarkers, thereby enabling more comprehensive in-silico cross-sectional and longitudinal analysis.

With the advancement of generative deep learning technologies, image synthesis has seen significant progress in recent years \cite{jang2023taupetgen,ou2024image,chen2025plasma}. Diffusion models, such as denoising diffusion probabilistic models (DDPM) \cite{ho2020denoising}, denoising diffusion implicit models (DDIM)\cite{song2020denoising}, and latent diffusion models (LDMs)\cite{rombach2022high}, have proved to have a high performance in PET image denoising\cite{gong2024pet,jiang2023pet,xie2024dose,shen2024bidirectional} and reconstruction \cite{singh2023score,webber2025supervised,phung2025joint,hashimoto2025pet}. Regarding synthetic medical image generation, recent text-guided diffusion models\cite{luo2024measurement} have shown outstanding performance in producing high-fidelity images under appropriate conditions. 
For example, a hierarchical diffusion-based framework was proposed to generate high-resolution lung CT images from radiology reports, demonstrating the feasibility of text-guided synthesis with anatomical consistency in volumetric data \cite{xu2024medsyn}. Rl4med-ddpo, a text-guided, latent diffusion model, was proposed and is capable of generating synthetic images from 6 medical specialties and 10 image types \cite{saremi2025rl4med}. For tau PET synthesis, our prior feasibility study, TauPETGen \cite{jang2023taupetgen}, shows that diffusion models can generate tau PET images conditioned on Mini-Mental State Examination (MMSE) scores. However, this approach is limited to generating 2D slices, restricting its applicability given PET’s inherently 3D nature.
\begin{figure*}[!t]
\centering
\includegraphics[width=0.95\textwidth]{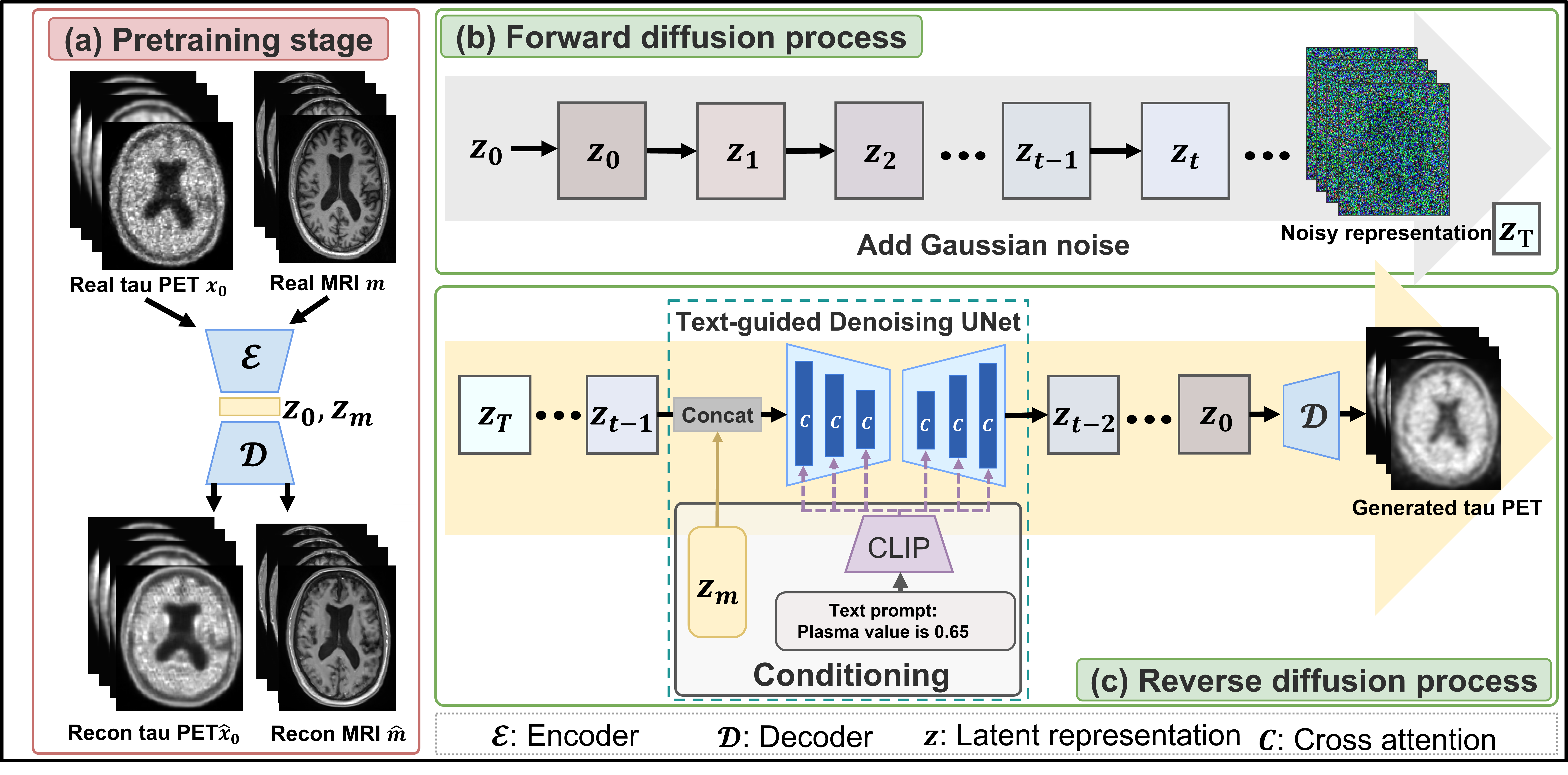} %Figure name
\caption{Overview of the proposed 3D TauGenNet framework: (a) Pretraining a 3D autoencoder for tau PET and MRI to obtain latent representations. (b) Forward diffusion process corrupting tau PET latent. (c) Reverse diffusion process during inference time to generate tau PET from the noisy latent. The text-guided denoising Unet accepts MR latent representation and text prompts as conditioning. }
\label{fig:figure1}
\end{figure*}

Advances in sensitivity and specificity for detecting tau pathology through plasma have established blood-based biomarkers as scalable tools for patient assessment, clinical trial recruitment, and disease monitoring in AD, providing a minimally invasive and cost-effective alternative to PET imaging. Among blood tau biomarkers, phosphorylated tau at threonine 217 (p-tau217) has shown high performance in differentiating AD from other neurodegenerative diseases and closely reflects cortical tau burden measured by tau-PET, making it a promising surrogate biomarker \cite{palmqvist2020discriminative,thijssen2021plasma,mielke2022performance,palmqvist2025plasma}. Given the strong correlation between plasma p-tau 217 and tau PET, plasma-based tau measures may serve as valuable predictors when estimating or modeling tau PET images. Recently, the ADNI database has released the p-tau 217 measures for the study participants. 

Building on recently released p-tau 217 data, we propose a 3D tau PET image generation framework using 3D latent diffusion models conditioned on both p-tau 217 measures and structural MR images. This design leverages MRI-derived anatomical priors while incorporating plasma biomarkers as conditioning inputs, resulting in more biologically informed and individualized tau PET synthesis. Beyond augmenting tau PET datasets, the framework has the potential to simulate disease progression by modulating plasma biomarker levels, thereby supporting cross-sectional and longitudinal analyses under diverse pathological conditions. The key contributions are as follows:

\begin{enumerate}
    \item This is the first work aiming to utilize p-tau 217 measures to aid tau PET image synthesis.  
    \item Considering that clinical PET images are acquired in a fully 3D mode, a 3D diffusion model-based synthetic image generation framework is developed. 
    \item We utilize CLIP to encode the p-tau 217 value through a text-based prompt, which is then combined with the MRI as an anatomical prior. This versatile integration enhances the quality of images generated across varying plasma levels and has the potential to further integrate clinical information from other unstructured resources.
    \item Our preliminary results suggest that the generated tau-PET images capture disease progression patterns that are consistent with the known trajectory of AD pathology.
\end{enumerate}

% while constructing a plasma-based prompt and extracting CLIP features to guide the U-Net of the latent diffusion model. 
\section{Method}
% Unlike conventional generative methods, our framework explicitly integrates structural magnetic resonance imaging (MRI) information and clinical plasma biomarker measurements (plasma p-tau217 levels) as complementary conditioning modalities. This design ensures that the generated tau PET images are both anatomically plausible and clinically meaningful, reflecting the subject’s biomarker status. 

In this work, we propose a multimodal conditional diffusion model to synthesize 3D tau PET images. T1-weighted MR images and p-tau 217 measures are utilized as the conditional input. The overall pipeline consists of three main stages: pretraining with a 3D autoencoder to learn latent representations, a forward diffusion process to gradually corrupt tau PET latents with noise, and a reverse denoising diffusion process during inference to generate the final image. Figure~\ref{fig:figure1} shows the diagram of the proposed framework.

\subsection{Autoencoder Pretraining Stage}
We first pretrain a 3D autoencoder to learn a compact latent representation from tau PET and MRI volumes. Although the training set includes both modalities, the autoencoder processes single-channel inputs, with PET and MRI images sharing the same network to minimize parameters. Let the ground-truth tau PET volume be denoted as $\mathbf{x}_0$, and the MRI volume be represented as $\mathbf{m}$. $\mathbf{x}_0$ and $\mathbf{m}$ have the same size and are registered. The tau PET latent representation can be obtained through the encoder $E$ as
\begin{equation}
\mathbf{z}_0 = E(\mathbf{x}_0),
\end{equation}
and the MRI latent representation can be obtained through the same encoder $E$ as
\begin{equation}
\mathbf{z}_m = E(\mathbf{m}).
\end{equation}
Here, $\mathbf{z}_0$ denotes the latent representation of the tau PET image, which is progressively perturbed with Gaussian noise during the forward diffusion process. $\mathbf{z}_m$ is the latent representation of the MR image, which remains uncorrupted and serves as structural guidance for the denoising network. The decoder $D(\cdot)$ tries to reconstruct the input image from its latent representation as
\begin{equation}
\hat{\mathbf{x}}_0 = D(\mathbf{z}_0), \quad \hat{\mathbf{m}} = D(\mathbf{z}_m).
\end{equation}

\subsection{Forward Diffusion Process}

In the forward diffusion process, only the tau PET latent $\mathbf{z}_0$ is progressively corrupted with Gaussian noise over $T$ discrete timesteps. The noise schedule is defined by variance parameters $\{\beta_t\}_{t=1}^T$, where $\beta_t \in (0,1)$. The cumulative noise parameters are
\begin{equation}
\alpha_t = 1 - \beta_t, \quad \bar{\alpha}_t = \prod_{s=1}^t \alpha_s.
\end{equation}
At timestep $t$, the noisy latent variable $\mathbf{z}_t$ is obtained as
\begin{equation}
\mathbf{z}_t = \sqrt{\bar{\alpha}_t} \mathbf{z}_0 + \sqrt{1 - \bar{\alpha}_t} \boldsymbol{\epsilon}, \quad \boldsymbol{\epsilon} \sim \mathcal{N}(\mathbf{0}, \mathbf{I}).
\end{equation}
As $t \to T$, $\mathbf{z}_t$ converges to isotropic Gaussian noise, $\mathbf{z}_T \sim \mathcal{N}(\mathbf{0}, \mathbf{I})$, which serves as the starting point for the reverse denoising process.
\begin{table*}[!t]
\centering
\caption{Mean Squared Error (MSE) Comparison Across Brain Regions for Plasma-Only and MRI+Plasma Conditions}
\label{tab:mse_comparison}
\begin{tabular}{lcccccc}
\hline
 & Parahippocampel & Fusiform & Inferior Temporal & Hippocampus & Posterior Cingulate & Entorhinal \\
\hline
Plasma & 0.028964 & 0.049624 & 0.235276 & 0.223135 & 0.212437 & 0.265567 \\
Plasma+MRI & 0.008911 & 0.008611 & 0.034286 & 0.014569 & 0.010041 & 0.025691 \\
\hline
\end{tabular}
\end{table*}
\subsection{Multi-Modal Conditioning}
To guide the diffusion process, the noise-free MRI latent $\mathbf{z}_m$ is concatenated with the noisy tau PET latent $\mathbf{z}_t$ along the channel dimension as
\begin{equation}
\mathbf{h}_t = [\mathbf{z}_t, \mathbf{z}_m].
\end{equation}
In addition, the plasma p-tau 217 value $v$ is converted into a natural language prompt and processed through a pretrained CLIP text encoder $\Phi_{\text{text}}(\cdot)$ to obtain a dense semantic embedding
\begin{equation}
\mathbf{c} = \Phi_{\text{text}}(\text{prompt}).
\end{equation}
This embedding encodes clinical biomarker information, enabling the diffusion process to generate PET images that reflect biologically relevant changes associated with different plasma p-tau 217 levels.
\subsection{Cross-Attention for Multimodal Conditioning}
The deployed U-Net employs layer-wise cross-attention modules, shown in Figure~\ref{fig:figure1} as 'C' at each layer. Suppose the U-Net has $L$ layers, and denote the feature map at layer $l$ as $\mathbf{F}_l$ with shape $(C,H,W,D)$. We first flatten the spatial dimensions into tokens
\begin{equation}
\mathbf{F}_{l,\text{tokens}} = \text{flatten}(\mathbf{F}_l),
\end{equation}
where $\mathbf{F}_{l,\text{tokens}}$ has shape $(N,C)$ with $N = H \cdot W \cdot D$.
Let the CLIP-encoded prompt embedding be $\mathbf{c}$ with shape $(M,D_c)$. For each layer $l$, we can project $\mathbf{c}$ into keys $\mathbf{K}_l$ and values $\mathbf{V}_l$ as
\begin{equation}
\mathbf{K}_l = \mathbf{W}_{k,l} \mathbf{c}, \quad \mathbf{V}_l = \mathbf{W}_{v,l} \mathbf{c},
\end{equation}
where $\mathbf{W}_{k,l}$ and $\mathbf{W}_{v,l}$ are learnable projection matrices, and $\mathbf{K}_l, \mathbf{V}_l$ have shapes $(M,d)$. We also project the feature tokens into queries $\mathbf{Q}_l$ as
\begin{equation}
\mathbf{Q}_l = \mathbf{W}_{q,l} \mathbf{F}_{l,\text{tokens}},
\end{equation}
where $\mathbf{W}_{q,l}$ is learnable and $\mathbf{Q}_l$ has the shape of $(N,d)$. Correspondingly, the attention weights and cross-attention output can be calculated as
\begin{equation}
\mathbf{A}_l = \text{softmax}\left(\frac{\mathbf{Q}_l \mathbf{K}_l^\top}{\sqrt{d}}\right), \quad \mathbf{O}_l = \mathbf{A}_l \mathbf{V}_l.
\end{equation}
Finally, we reshape $\mathbf{O}_l$ back to the original feature map shape and add a residual connection as
\begin{equation}
\mathbf{F}_l' = \mathbf{F}_l + \text{reshape}(\mathbf{O}_l).
\end{equation}
This layer-wise cross-attention is applied to every layer $l = 1,\dots, L$ of the U-Net. This allows the CLIP prompt embedding $\mathbf{c}$ to modulate features at multiple resolutions, conditioning tau PET generation on plasma biomarker information across scales.
\subsection{Reverse Diffusion Process}
The reverse diffusion process is parameterized by a denoising U-Net $\epsilon_\theta$, which learns to iteratively remove noise and recover the clean tau PET latent. At each timestep $t$, the network takes the concatenated latent $\mathbf{h}_t$, the current timestep $t$, and the clinical embedding $\mathbf{c}$, and predicts the noise component:
\begin{equation}
\hat{\boldsymbol{\epsilon}}_\theta = \epsilon_\theta(\mathbf{h}_t, t, \mathbf{c}).
\end{equation}
The model is trained using the standard diffusion loss, minimizing the mean squared error between the true Gaussian noise $\boldsymbol{\epsilon}$ and the predicted noise $\hat{\boldsymbol{\epsilon}}_\theta$:
\begin{equation}
\mathcal{L}(\theta) = \mathbb{E}_{t \sim \mathcal{U}(1,T), \boldsymbol{\epsilon} \sim \mathcal{N}(\mathbf{0},\mathbf{I})} \left[ \|\boldsymbol{\epsilon} - \epsilon_\theta(\mathbf{h}_t, t, \mathbf{c})\|_2^2 \right].
\end{equation}
During inference, generation starts from pure Gaussian noise $\mathbf{z}_T \sim \mathcal{N}(\mathbf{0}, \mathbf{I})$, and the model iteratively applies the learned denoising steps to recover the clean latent $\mathbf{z}_0$. The final synthesized tau PET volume is obtained by decoding the recovered latent $\hat{\mathbf{z}}_0$ as 
\begin{equation}
\hat{\mathbf{x}}_0 = D(\hat{\mathbf{z}}_0).
\end{equation}
Through this conditioning strategy and layer-wise cross-attention, the model generates tau PET volumes that are anatomically aligned with MRI structure and modulated according to plasma biomarker status, ensuring both anatomical fidelity and clinical interpretability.
\section{EXPERIMENTS}
\subsection{Dataset and Implementation Details} 
A total of 360 paired 3D tau PET and MRI images from the ADNI database \cite{mueller2008alzheimer} were utilized in this study. The tau PET image was scaled to have the unit of SUVR. All images were spatially aligned and resampled to a fixed resolution of $160 \times 160 \times 96$ voxels and normalized to the range $[0,1]$. Each tau PET-MRI pair was accompanied by a plasma p-tau 217 value. The p-tau 217 value was embedded into natural language prompts in the format
``\texttt{Plasma is [p-tau 217 value].}"
 to enable text-conditioned image synthesis. Among the 360 paired datasets, 260 pairs were used for training, and 100 pairs were held out for testing. To ensure spatial correspondence between PET and MRI, we applied rigid registration using ANTs \cite{avants2011reproducible}. Specifically, the MRI volumes were used as moving images and the corresponding PET volumes as fixed images. The registration aligns the overall position and orientation of MRI to PET, and the transformed MRI volumes were saved for subsequent 3D tau PET image synthesis. Freesurfer \cite{desikan2006automated} was utilized to get the region of interest and construct the surface maps based on the T1-weighted MR images. 
 All experiments were implemented in PyTorch v1.12 and ran on a single NVIDIA A100 GPU (80 GB memory). The model was trained with the Adam optimizer using a learning rate of $1\times10^{-4}$ and a batch size of 8, with an epoch number of 600.  
 
\begin{figure*}[!t]
\centering
\includegraphics[width=\textwidth]{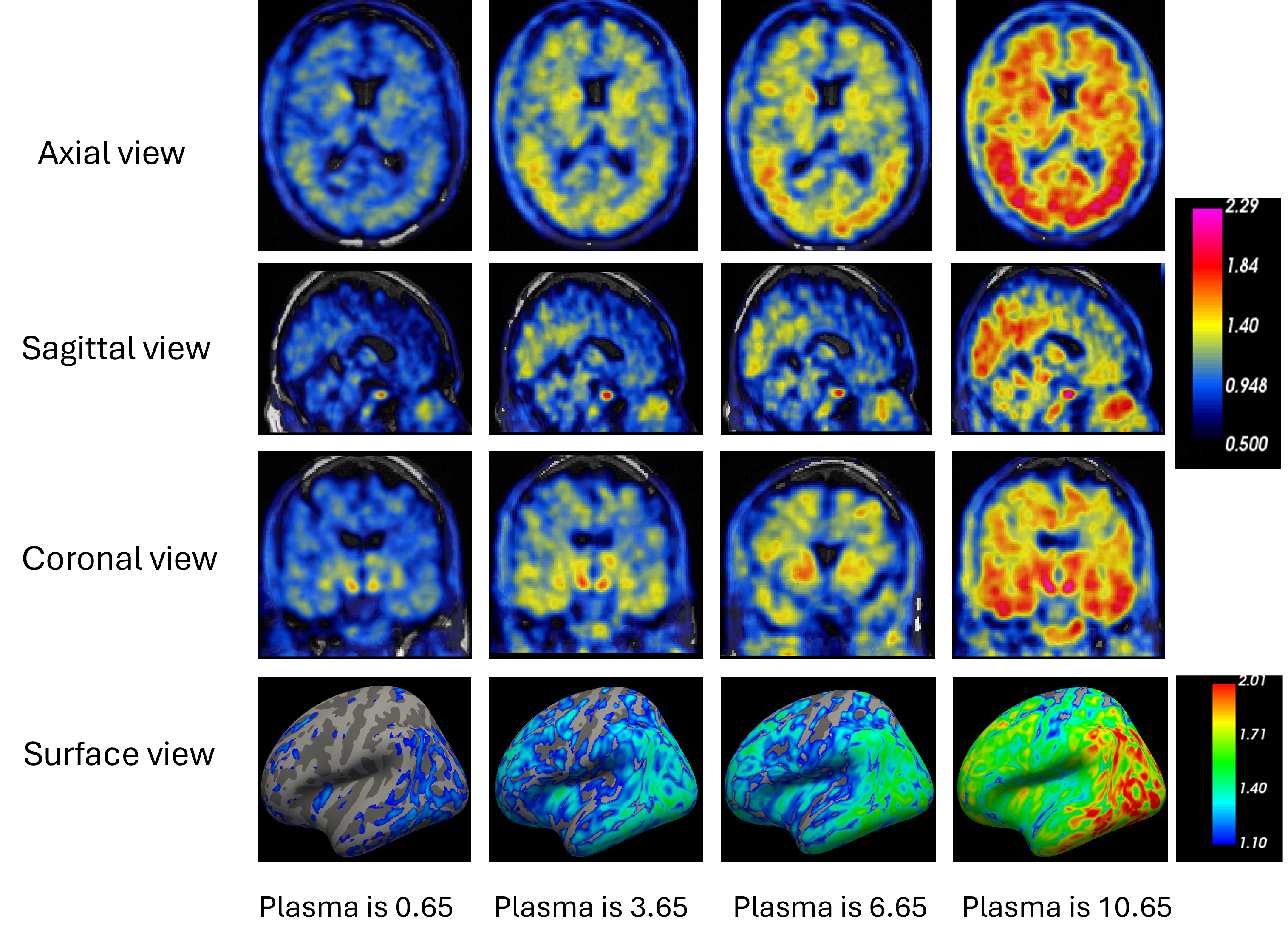} 
\caption{One example of generated tau PET images across varying p-tau 217 concentrations (0.65, 3.65, 6.65, and 10.65) shown at axial, sagittal, and coronal views (top three rows). The last rows show the surface images at the corresponding p-tau 217 values.}
\label{fig:Figure4}
\end{figure*}

\begin{table*}[!t]
\centering
\caption{Region-wise mean squared error (MSE) of synthetic tau-PET stratified by plasma p-tau 217 intervals.}
\label{tab:mse_tau_pet}
\begin{tabular}{lcccccc}
\hline
% \textbf{Plasma bin} & \textbf{Parahippocampal} & \textbf{Fusiform} & \textbf{Inferior Temporal} & \textbf{Hippocampus} & \textbf{Posterior Cingulate} & \textbf{Entorhinal} & & & & & & \\
\textbf{Plasma bin} & \textbf{Parahippocampal} & \textbf{Fusiform} & \textbf{Inferior Temporal} & \textbf{Hippocampus} & \textbf{Posterior Cingulate} & \textbf{Entorhinal} \\
\hline
0-2 & 0.019792 & 0.033371 & 0.012390 & 0.040414 & 0.021962 & 0.035778 \\
2-4 & 0.000143 & 0.000808 & 0.003748 & 0.001388 & 0.001104 & 0.004394 \\
4-6 & 0.016721 & 0.049564 & 0.011945 & 0.006732 & 0.007424 & 0.051724 \\
6-8 & 0.005301 & 0.016692 & 0.004161 & 0.002042 & 0.000035 & 0.031331 \\
10+ & 0.000596 & 0.147683 & 0.141186 & 0.021271 & 0.003955 & 0.005230 \\
\hline
\end{tabular}
\end{table*}

\begin{figure*}[!t]
\centering
\includegraphics[width=0.85\textwidth]{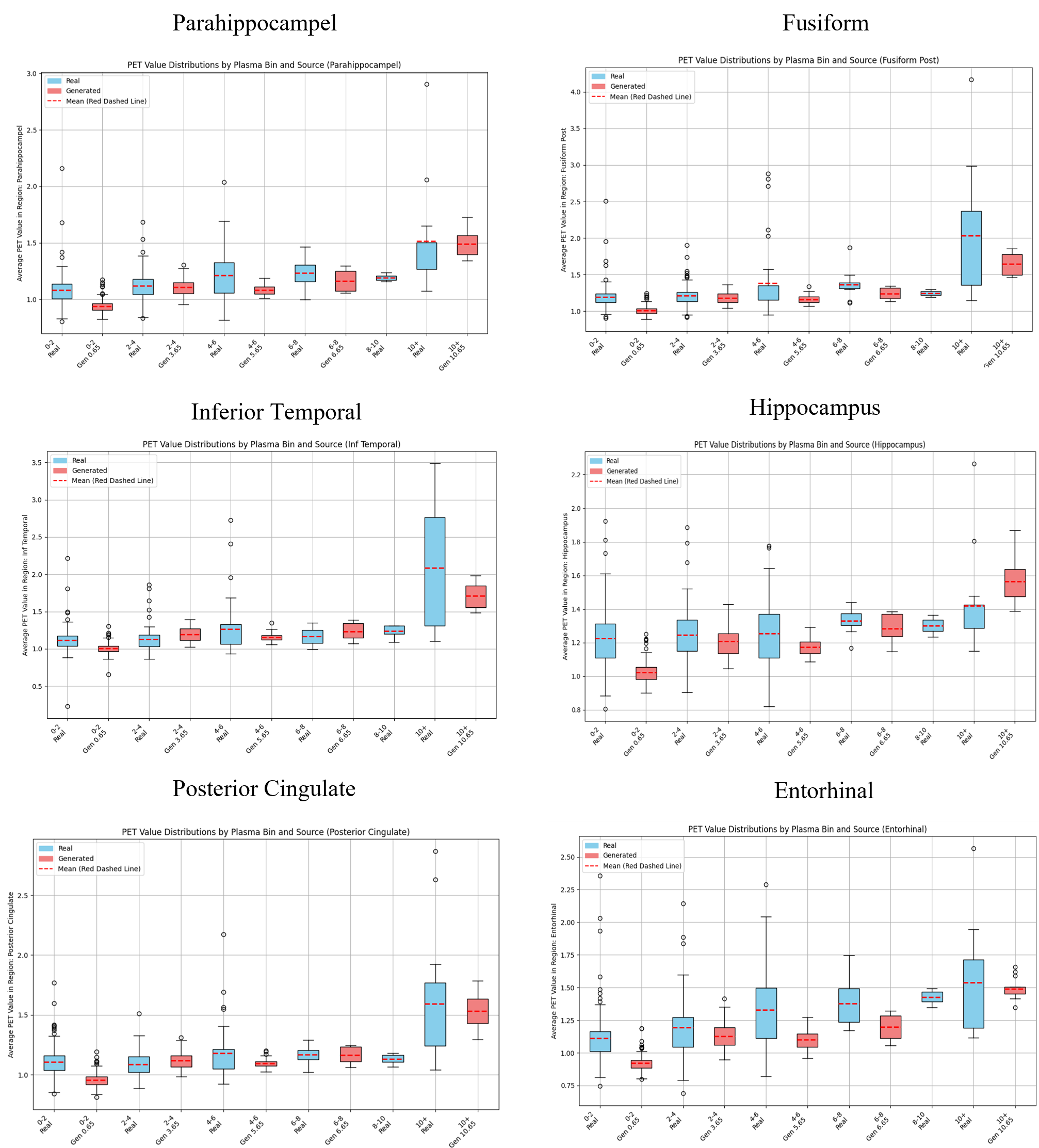} 
\caption{The box plot of the average values of different brain regions for the generated and the real measured tau images at different p-tau 217 ranges.}
\label{fig:figure2}
\end{figure*}

\begin{figure*}[!t]
\centering
\includegraphics[trim=0.3cm 0.5cm 0cm 0cm, clip,width=0.95\textwidth]{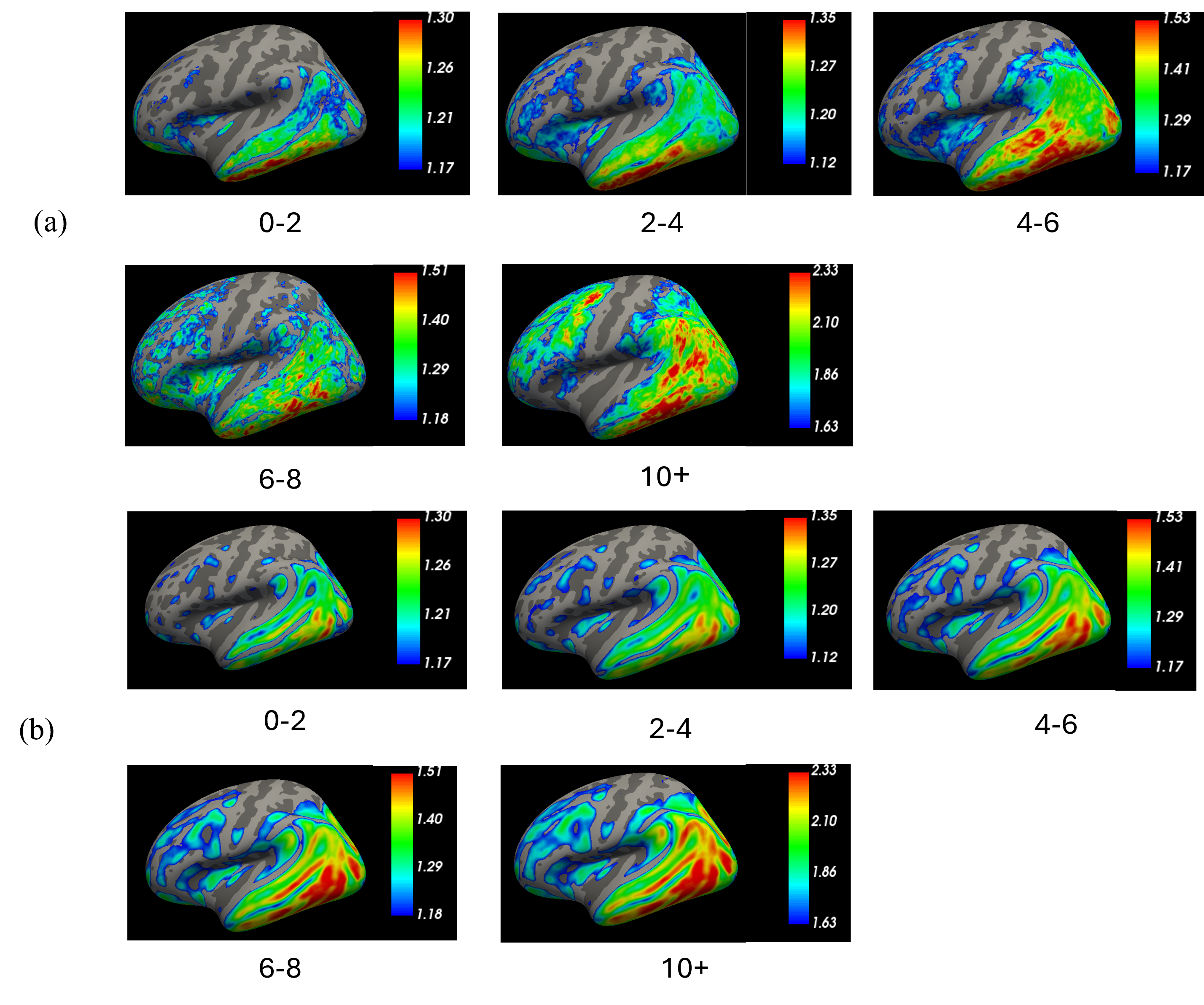} %Figure name
\caption{The brain surface plots under different p-tau 217 ranges for (a) real tau PET images and (b) generated tau PET images.}
\label{fig:figure3}
\end{figure*}

\subsection{Evaluation}

To quantitatively evaluate the quality of the generated tau PET images across different plasma intervals and multiple tau progression-related brain regions, we first compute the mean tau PET value within each brain region for each individual sample. Specifically, let $N_g$ denote the number of samples in the $g$-th plasma interval, and $R_{i,g,k}^{\text{real}}$ and $R_{i,g,k}^{\text{gen}}$ represent the mean tau PET values of the $i$-th sample in the $k$-th brain region for the real and generated images, respectively. Next, we calculate the group-level mean for each brain region by averaging these individual means across all samples within the plasma interval:
\begin{align}
\bar{R}_{g,k}^{\text{real}} &= \frac{1}{N_g} \sum_{i=1}^{N_g} R_{i,g,k}^{\text{real}}, \\
\bar{R}_{g,k}^{\text{gen}} &= \frac{1}{N_g} \sum_{i=1}^{N_g} R_{i,g,k}^{\text{gen}}.
\end{align}
We then define the mean squared error (MSE) for each plasma interval and brain region as the squared difference between the real and generated group means as
\begin{equation}
\text{MSE}_{g,k} = \left( \bar{R}_{g,k}^{\text{real}} - \bar{R}_{g,k}^{\text{gen}} \right)^2.
\end{equation}
This procedure results in an error matrix of size $G \times K$, where each element reflects the discrepancy between the generated and real PET images in a specific plasma interval and brain region at the group average level. This evaluation can characterize the model’s generative performance across multiple AD-related brain regions and biomarker value ranges.
% , while preserving fine-grained variations that would be lost in a single aggregated metric.

\section{Result}

\subsection{Benefits of utilizing both MR and plasma as input}

TABLE~\ref{tab:mse_comparison} shows that when tau PET images were generated using plasma values alone (plasma-only model), the prediction errors were consistently higher across all brain regions, particularly in regions strongly associated with tau deposition, such as the entorhinal cortex (0.265567) and the hippocampus (0.223135). This indicates that plasma values alone are insufficient to capture the spatial distribution patterns of tau PET. In contrast, incorporating MRI as an additional structural constraint (MRI + plasma model) led to a substantial reduction in errors across all regions. For instance, the errors in the  entorhinal cortex and hippocampus decreased to 0.025691 and 0.014569, respectively. These results suggest that structural information from MRI effectively guides the model to generate tau PET images that better match the true spatial patterns, thereby markedly improving prediction accuracy.
\subsection{Visual examples}
Figure~\ref{fig:Figure4} shows an example of the generated tau PET images across varying p-tau 217 values (0.65 to 10.65) while maintaining fixed structural MRI information. The visualization shows that the generated tau PET images exhibit a progressive increase in tracer uptake intensity and spatial extent with increasing plasma biomarker concentrations. At lower plasma values, tau signals are predominantly localized to early tau-related structures, while higher concentrations show expansion into widespread cortical regions. This progression pattern aligns with established tau pathology spread in AD, suggesting the proposed generative model might be able to capture the spatial-temporal relationship between plasma biomarkers and brain pathology. 

\subsection{Prediction error across various plasma ranges}
TABLE~\ref{tab:mse_tau_pet} presents the MSE values between the predicted and ground-truth tau PET region-average values across six AD-related brain regions, grouped by plasma level bins. In the 0--2 plasma range, moderate MSE values are observed, indicating reasonable agreement between generated and real tau distributions in early-stage subjects. The lowest MSE is consistently found in the 2--4 bin across all regions (e.g., 0.0001 in the parahippocampal gyrus and 0.0008 in the fusiform cortex), suggesting that the model performs best when plasma levels are within a stable early-to-mid pathological range. In the 4--6 range, error increases in several regions, notably the fusiform cortex (0.0496) and entorhinal cortex (0.0517), which are among the earliest sites of tau accumulation in AD. This trend continues in the 6--8 bin. In the 10+ plasma group, MSE values increase substantially in regions such as the fusiform cortex (0.1477) and inferior temporal lobe (0.1412). These results indicate greater prediction difficulty in these regions, likely due to increased pathological variability when the p-tau 217 levels are high. Overall, the predicted region- and stage-specific error patterns align well with the measured data, suggesting that the proposed model captures biologically meaningful changes conditioned on plasma levels.

\subsection{Group-level comparison between generated and ground-truth tau images}
As shown in Fig.~\ref{fig:figure2}, the boxplot analysis across stratified plasma tau intervals demonstrates clear interval-specific patterns and strong biological concordance in the generative model’s ability to replicate the distributional characteristics of real PET signals. Within the baseline p-tau 217 range (0--2), all six brain regions exhibit exceptional concordance between generated and real values, establishing a robust foundation for model reliability. As plasma tau values increased to the 2--4 interval, the generative model reached its optimal performance range, which is clinically significant as this interval corresponds to tau burden levels typical of mild cognitive impairment and early dementia. The parahippocampal and fusiform regions exhibit near-perfect concordance, with boxplots of generated and real values nearly overlapping completely. The entorhinal cortex, as the earliest affected region in AD, demonstrate stable generative performance within this interval, with outlier distributions closely matching real data. For plasma tau intervals from 4 to 10, the generated-image plots and the real-data plots overall matched well. Across almost all regions and plasma levels, the variance of the generated data is lower than that of the real measurements, with the discrepancy most evident at plasma tau values greater than 10. A likely explanation is that our model is currently limited to p-tau 217 changes, while real data additionally capture variability stemming from demographic and clinical factors. By including such variables as model inputs, we anticipate an increase in variance that more closely aligns with the observed data.

\subsection{Surface analysis}
Figure~\ref{fig:figure3} illustrates the average cortical surface distribution of tau PET accumulation in both observed and generated PET images across different plasma p-tau217 intervals (0--2, 2--4, 4--6, 6--8, and 10+). Both sets of images represent the mean tau signal on the cortical surface derived from multiple subjects within each plasma range. As shown in Figure~\ref{fig:figure3}(a), based on real measured tau PET data, increasing plasma p-tau 217 levels corresponds to a progressive intensification of tau signal, initially localized at the entorhinal cortex with low intensity, and subsequently expanding to the temporal lobe, posterior cingulate cortex, and parietal regions, reflecting the canonical spatial progression pattern of tau pathology in AD. The generated PET images demonstrate a consistent dynamic pattern across the plasma intervals: at low plasma levels, tau signal is weak and predominantly confined to the entorhinal cortex; at moderate levels, signal intensity increases and extends into the temporal lobe; at higher levels, tau signal markedly intensifies and spreads to the posterior cingulate and parietal cortices; and at the highest plasma range, a widespread and robust tau distribution is observed. These experiments indicate that the proposed network accurately captures and reproduces the spatial distribution characteristics of tau pathology across plasma p-tau217 ranges, validating its biological plausibility and sensitivity to plasma changes. This establishes a solid foundation for future investigations employing the model in pathological progression simulation and biomarker association studies.~\cite{johnson2016tau}.
\section{Discussion}

In this work, we propose a framework that integrates structural MR images with plasma p-tau 217 values to generate individualized tau PET images without requiring actual tau PET scans, which remain limited in availability. This approach extends the utility of blood-based biomarkers for characterizing AD progression and offers several practical applications. First, it enables data augmentation to expand the availability of tau PET data, which is valuable in missing-data settings and for machine learning tasks such as large-scale pretraining. Second, by simulating tau PET changes under varying plasma levels, the framework provides a tool for in silico analyses of disease trajectories and therapeutic response. One thing to note is that we do not claim that the generated tau PET images are suitable for clinical diagnosis. Tau PET has a substantially higher dynamic range compared to 18F-FDG or amyloid PET, which increases the risk of prediction error. Further model refinement and incorporation of additional clinical information will be necessary to enhance the diagnostic value of these synthetic tau PET images.

This study has several limitations. First, the relationship between plasma biomarkers and tau PET is inherently complex, and the relatively small dataset (360 paired samples) limits the model’s ability to learn stable and accurate mappings, potentially reducing the reliability of the generated images. Future work will address this by incorporating larger datasets of tau PET images and plasma measures from multiple sources. Second, the present framework is restricted to p-tau 217–driven tau PET synthesis and does not yet integrate other biomarkers. To enhance biological plausibility, our future studies will potentially incorporate longitudinal tau PET data, amyloid PET data, additional plasma biomarkers, demographic information, and cognitive measures (e.g., MMSE). These complementary inputs can be integrated into the framework via the CLIP model and image encoders.

\section{Conclusion}
In this work, we propose a plasma- and MRI-informed 3D diffusion model for synthetic tau PET image generation that captures biologically relevant patterns of tau deposition. Experimental results show that the generated images not only appear realistic but also exhibit spatial distributions consistent with AD progression. When plasma levels vary, the synthetic tau PET images demonstrate region-specific changes that align with established biological trends in clinical data. The framework can support data augmentation of tau PET datasets and enable in silico analyses of disease trajectories and therapeutic response under different p-tau 217 conditions. Future work will focus on scaling the model with larger training cohorts and incorporating additional clinical and biomarker information to further enhance its reliability and biological plausibility.

\section*{Acknowledgment}
Data collection and sharing for this work was funded by the Alzheimer's Disease Neuroimaging Initiative (ADNI) 
(NIH Grant U01 AG024904; Department of Defense ADNI W81XWH-12-2-0012). 
ADNI is funded by the National Institute on Aging, the National Institute of Biomedical Imaging and Bioengineering, 
and through contributions from numerous public and private partners. 
The investigators within ADNI contributed to the design and implementation of ADNI and/or provided data but did not participate 
in analysis or writing of this report. 
This study used ADNI 1234GO PET and MRI imaging data, along with associated clinical and biomarker information provided in ADNI Excel files.

\balance
\bibliographystyle{ieeetr}  
\bibliography{references}   
\end{document}